\documentclass[10pt,twocolumn,letterpaper]{article}

\usepackage{cvpr}              

\usepackage{multicol}
\usepackage{multirow}
\usepackage{graphicx,verbatim}
\usepackage{booktabs}
\usepackage{amsmath,amssymb}
\usepackage{multirow}
\usepackage{array}
\usepackage{bm}

\definecolor{cvprblue}{rgb}{0.21,0.49,0.74}
\usepackage[pagebackref,breaklinks,colorlinks,allcolors=cvprblue]{hyperref}

\def\paperID{*****} 
\def\confName{CVPR}
\def\confYear{2026}

\title{Instance Awareness of Multi-class Semantic Segmentation Loss Functions}

\author{Soumya Snigdha Kundu$^{1}$ \quad Florian Kofler$^{2}$ \quad Marina Ivory$^{1}$ \quad Hendrik M\"oller$^{3}$\\
Jonathan Shapey$^{1,4}$ \quad Tom Vercauteren$^{1}$\\[0.5em]
$^{1}$King's College London \quad
$^{2}$University of T\"ubingen \quad
$^{3}$Technical University of Munich\\
$^{4}$King's College Hospital\\[0.3em]
{\tt\small \{soumya\_snigdha.kundu, marina.ivory, jonathan.shapey, tom.vercauteren\}@kcl.ac.uk}\\
{\tt\small florian.kofler@uni-tuebingen.de \quad hendrik.moeller@tum.de}
}

\begin{document}
\maketitle

\begin{abstract}
	Instance-sensitive losses for semantic segmentation such as blob loss and CC loss were designed to address \emph{instance imbalance}, ensuring small lesions generate the same gradient as large ones, but operate only on single-class segmentation.
	In multi-class settings, \emph{class imbalance} poses an additional problem: rare classes with few instances receive a disproportionately small share of the training signal.
	We show that extending instance-sensitive losses to multi-class segmentation via a one-vs-rest class decomposition repurposes them to also address class imbalance, as uniform averaging over classes ensures each class contributes equally regardless of frequency.
	We further show that inverse-size weighting, which destabilizes training when applied globally due to weight imbalances across rare and common classes, becomes effective when integrated within the per-component loss, confining the reweighting to each component's spatial context.
	On the BraTS-METS 2025 dataset (260 test cases), multi-class CC loss improves foreground Dice ($0.64 \pm 0.26$ vs.\ $0.59 \pm 0.27$ baseline) and rare-class Dice, while maintaining Panoptic Quality at DSC threshold 0.5. Multi-class blob loss achieves the best Panoptic Quality at threshold 0.5 ($0.40 \pm 0.24$ vs.\ $0.38 \pm 0.25$ baseline) and recognition quality ($0.53 \pm 0.29$ vs.\ $0.49 \pm 0.30$). Integrating inverse-size weighting within the per-component loss increases rare-class Dice to $0.44 \pm 0.36$ at the cost of reduced detection quality.
\end{abstract}

\begin{figure}[t]
	\centering
	\includegraphics[width=\columnwidth]{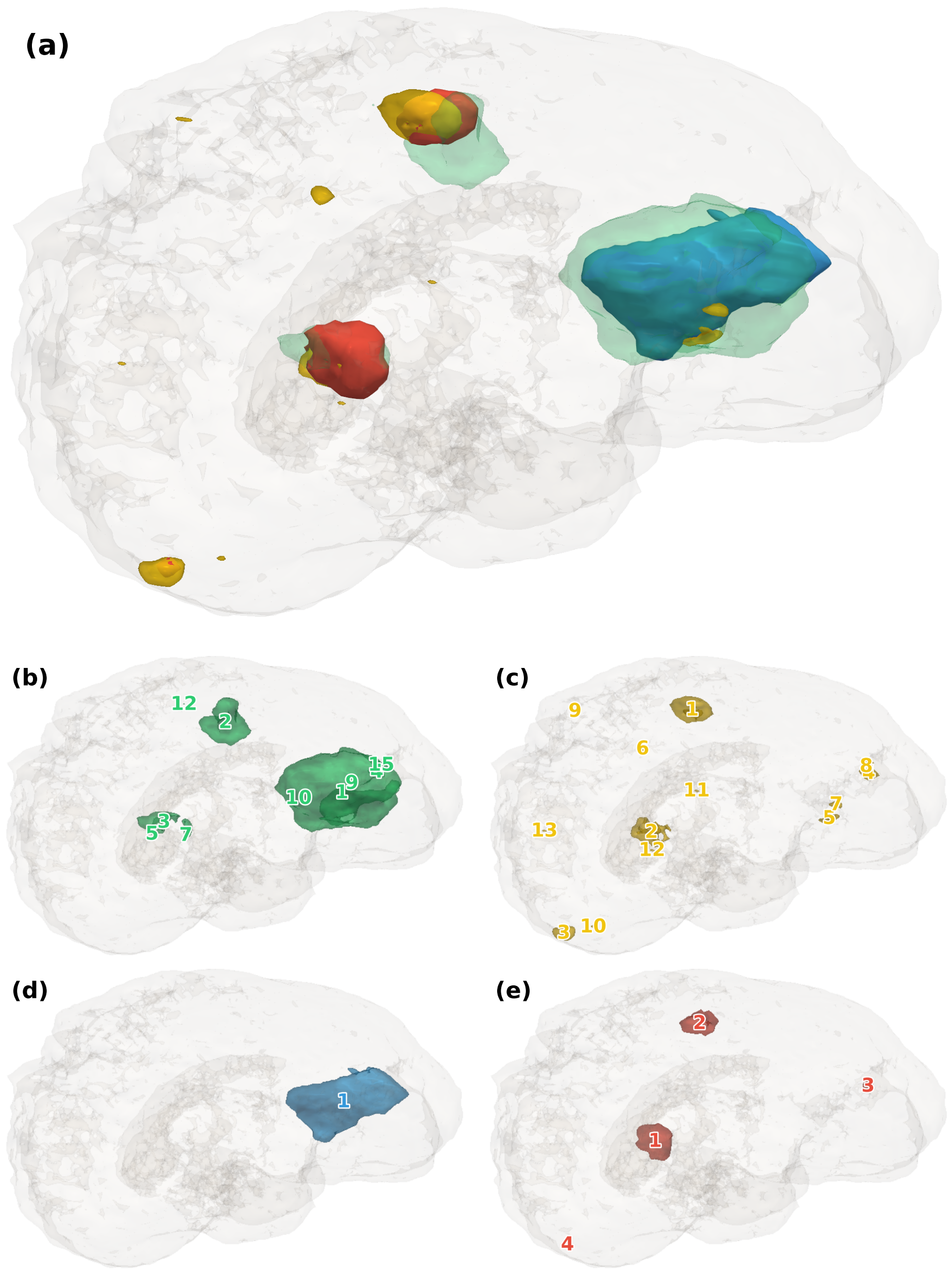}
	\caption{3D rendering of a representative BraTS-METS case. (a) Semantic reference label with four tumor sub-regions overlaid on the T1c brain volume. (b--e) Per-class instance decomposition via connected components: (b) \textcolor[HTML]{2ecc71}{SNFH} (15 components), (c) \textcolor[HTML]{f1c40f}{ET} (17 components), (d) \textcolor[HTML]{3498db}{RC} (1 component), (e) \textcolor[HTML]{e74c3c}{NETC} (4 components). Numbers index individual instances for each class.}\label{fig:dataset}
\end{figure}

\section{Introduction}

Brain metastases are the most common intracranial tumors, affecting 10--26\% of patients who die from cancer~\cite{nayak2012epidemiology}, with an estimated 98{,}000--170{,}000 new cases per year in the United States alone~\cite{smedby2009brain}.
Stereotactic radiosurgery (SRS) is a standard treatment for both intact and resected brain metastases, requiring precise target delineation to deliver tumoricidal doses while sparing surrounding tissue~\cite{minniti2021current}.
For resected metastases, adjuvant SRS to the resection cavity reduces 12-month local recurrence from approximately 50\% to under 30\%~\cite{mahajan2017post}, making accurate cavity segmentation directly relevant to treatment planning.
Individual metastases range from sub-millimeter micro-metastases to lesions exceeding several centimeters, and SRS dose prescription depends directly on target volume~\cite{minniti2021current}, making accurate size estimation essential.
Beyond SRS targeting, longitudinal monitoring of individual tumor sub-regions enables quantitative tracking across timepoints~\cite{maleki2025analysis}.

The BraTS-METS 2025 challenge formalizes this clinical need by defining four segmentation regions: whole tumor (WT), tumor core (TC), enhancing tumor (ET), and resection cavity (RC)~\cite{maleki2025analysis}.
Automated segmentation of these sub-regions is therefore essential for scalable, consistent treatment planning and follow-up, but standard segmentation networks trained with voxel-level losses such as Dice + cross-entropy (DC+CE) perform well on common structures and struggle with rare ones.
Multi-class Dice loss partially addresses class imbalance by computing a separate score per class and averaging uniformly, giving each class equal weight regardless of voxel count, but it still operates at the voxel level and does not distinguish between instances.

\begin{figure*}[t]
	\centering
	\includegraphics[width=\textwidth]{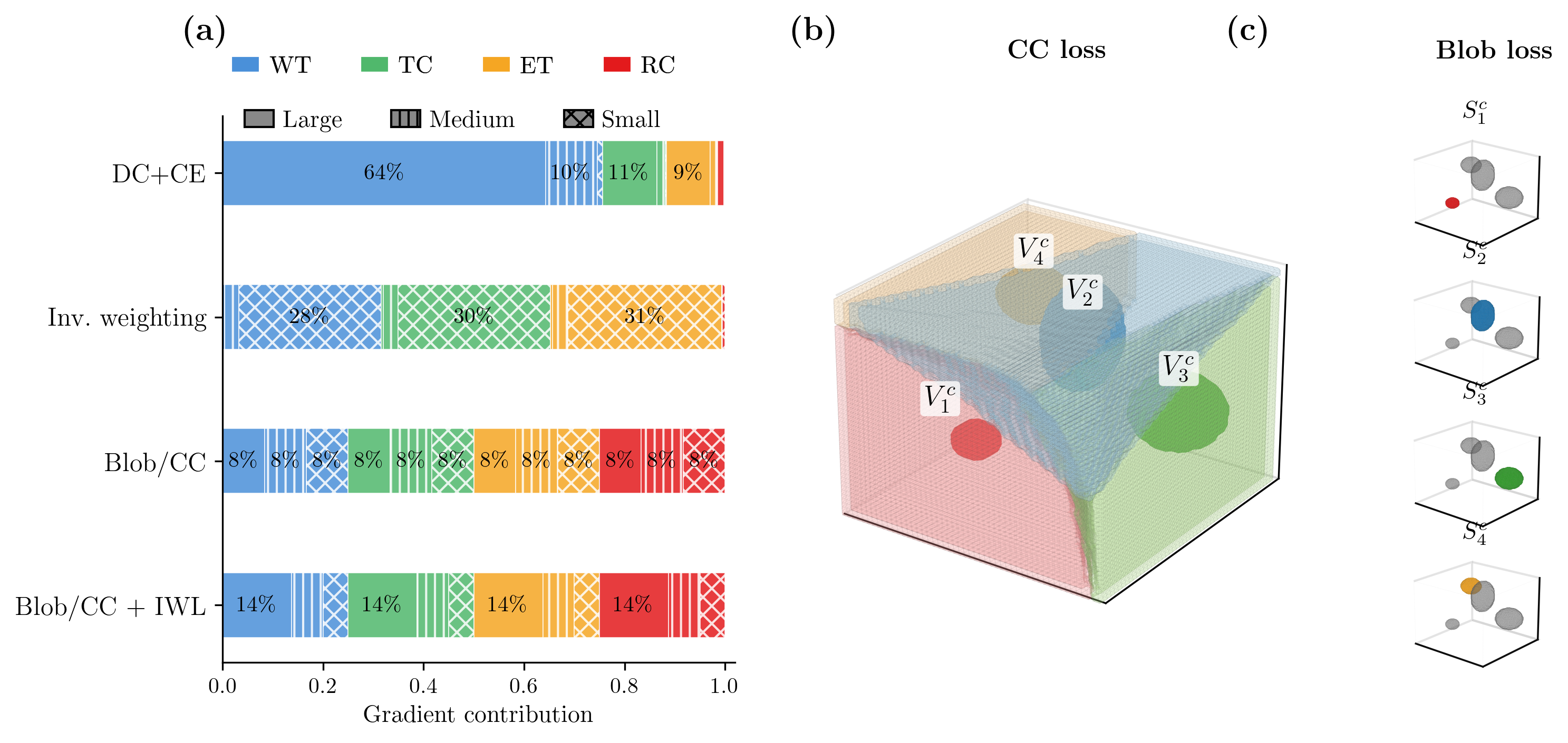}
	\caption{Overview of instance-sensitive loss decomposition. (a) Gradient contribution by class (color) and instance size (hatch). DC+CE is dominated by large instances in common classes; inverse-size weighting shifts to small instances but RC remains underrepresented; Blob/CC with class decomposition equalizes both classes and instances; adding IWL within components further emphasizes small instances while preserving class balance. (b) CC loss partitions the volume into Voronoi cells $V_k^c$. (c) Blob loss pairs each component $S_k^c$ with the background.}\label{fig:method}
\end{figure*}

On the BraTS-METS 2025 test set (a randomized split, as the original test set is not publicly available), resection cavity (RC) exhibits triple imbalance: present in only 10\% of cases, contributes 32 components versus 1{,}739 for ET, and represents 295K voxels versus 15.5M for whole tumor.
While Dice loss normalizes per class, the cross-entropy component averages gradients across all foreground voxels, so classes with fewer voxels receive proportionally smaller updates, compounding the effect of case- and instance-level rarity.
Under DC+CE training, the baseline nnU-Net~\cite{isensee2021} achieves an RC Dice Similarity Coefficient (DSC) of only 0.15 and a foreground-mean Panoptic Quality~\cite{kirillov2019} at DSC threshold 0.5 ($PQ_{0.5}^{\text{DSC}}$) of 0.38.

Instance-sensitive losses address a related but different problem: \emph{instance imbalance}, where small lesions are under-represented relative to large ones within the same class.
One of the earliest efforts was inverse-size weighting~\cite{shirokikh2020}, where each voxel is assigned a weight inversely proportional to its connected-component size, amplifying gradients from small instances.
More recently, Blob loss~\cite{kofler2023} and CC loss~\cite{bouteille2025learning} have taken this further by decomposing the label map into connected components and evaluating an independent loss per component, ensuring each instance contributes equally to the gradient. Connected components derived from the binary label map serve as a proxy for instances in the absence of explicit instance masks, allowing these losses to be instance-sensitive without requiring ground-truth instance annotations. However, all three methods were developed for single-class segmentation and equalize gradients only within each class, not across classes with vastly different frequencies.

To our knowledge, this work is among the first to address instance and class imbalance simultaneously within a single loss for multi-class segmentation.
We observe that extending instance-sensitive losses to multi-class segmentation, via a one-vs-rest decomposition that treats each foreground class independently, repurposes these losses to solve \emph{class imbalance} in addition to instance imbalance.
By averaging per-component losses uniformly across classes, instances from each class receive equal gradient weight regardless of how rare the class is (see Fig.~\ref{fig:method}).
A 78\,mm\textsuperscript{3} micro-metastasis in a common class and a 2{,}617\,mm\textsuperscript{3} resection cavity in a rare class contribute equally to the training signal.

We further integrate inverse-size weighting \emph{within} the per-component loss, rather than applying it as a standalone global mechanism, allowing it to interact with the component-level decomposition.
Our evaluation reveals a systematic trade-off between voxel-level overlap (DSC) and instance-level detection ($PQ_{0.5}^{\text{DSC}}$): methods that up-weight rare classes improve DSC at the cost of lower $PQ_{0.5}^{\text{DSC}}$.
Our contributions are as follows:
\begin{itemize}
	\item We show that instance-aware losses (blob loss, CC loss), when applied to multi-class segmentation via one-vs-rest class decomposition, are repurposed from solving instance imbalance to also solving class imbalance, because uniform averaging over classes ensures each class contributes equally regardless of frequency.
	\item We show that inverse-size weighting, which destabilizes training when applied globally to multi-class segmentation due to weight imbalances across rare and common classes (small instances in rare classes receive much higher weights than large instances in common classes), becomes effective when integrated within the per-component loss, increasing blob loss RC Dice from 0.22 to 0.44, as the reweighting is confined to each component's spatial context.
\end{itemize}

\begin{table*}[t]
\caption{Dataset statistics for BraTS-METS 2025 (260 test cases), illustrating the triple imbalance affecting RC: it is rare at the case, instance, and voxel levels.}\label{tab:dataset}
\centering
\begin{tabular}{lcccc}
\toprule
Region & Cases present & Components & Median size (mm\textsuperscript{3}) & Total voxels \\
\midrule
WT & 253 (97\%) & 1{,}605 & 120 & 15.5M \\
TC & 251 (97\%) & 1{,}715 & 82 & 2.6M \\
ET & 251 (97\%) & 1{,}739 & 78 & 2.1M \\
RC & 26 (10\%) & 32 & 2{,}617 & 295K \\
\bottomrule
\end{tabular}
\end{table*}

\begin{table*}[t]
	\caption{Size-stratified foreground-mean $PQ_{0.5}^{\text{DSC}}$ by instance size (small $<$0.5\,cc, medium 0.5--5\,cc, large $>$5\,cc). Best in \textbf{bold}, second-best \underline{underlined}. $\dagger$: $\beta = 2$.}\label{tab:size}
	\centering
	\begin{tabular}{lccc}
		\toprule
		\multirow{2}{*}{Method}      & \multicolumn{3}{c}{$PQ_{0.5}^{\text{DSC}}$}                                                                 \\
		\cmidrule(lr){2-4}
		                             & Small                                       & Medium                        & Large                         \\
		\midrule
		Baseline (DC+CE)             & $\underline{0.229 \pm 0.246}$               & $0.565 \pm 0.280$             & $\underline{0.712 \pm 0.210}$ \\
		\midrule
		Global + Blob                & $\bm{0.241 \pm 0.247}$                      & $0.585 \pm 0.254$             & $0.703 \pm 0.202$             \\
		Global + Blob$^\dagger$      & $0.223 \pm 0.236$                           & $0.562 \pm 0.256$             & $0.688 \pm 0.196$             \\
		Global + CC                  & $0.227 \pm 0.242$                           & $\underline{0.598 \pm 0.275}$ & $\bm{0.730 \pm 0.212}$        \\
		Global + CC$^\dagger$        & $0.229 \pm 0.243$                           & $\bm{0.611 \pm 0.258}$        & $0.710 \pm 0.209$             \\
		\midrule
		Global + IWL-Blob            & $0.153 \pm 0.191$                           & $0.520 \pm 0.299$             & $0.707 \pm 0.215$             \\
		Global + IWL-Blob$^\dagger$  & $0.111 \pm 0.151$                           & $0.468 \pm 0.325$             & $0.690 \pm 0.207$             \\
		Global + IWL-CC              & $0.122 \pm 0.169$                           & $0.500 \pm 0.317$             & $0.674 \pm 0.207$             \\
		Global + IWL-CC$^\dagger$    & $0.069 \pm 0.131$                           & $0.427 \pm 0.322$             & $0.620 \pm 0.222$             \\
		\midrule
		InvWeight (global)           & $0.041 \pm 0.096$                           & $0.345 \pm 0.316$             & $0.586 \pm 0.281$             \\
		InvWeight$^\dagger$ (global) & $0.017 \pm 0.036$                           & $0.290 \pm 0.294$             & $0.533 \pm 0.294$             \\
		InvWeight (local)            & $0.003 \pm 0.007$                           & $0.219 \pm 0.257$             & $0.425 \pm 0.326$             \\
		\bottomrule
	\end{tabular}
\end{table*}

\section{Methods}

We consider multi-class 3D segmentation with $C$ classes (including background).
Let $y \in \{0, \ldots, C{-}1\}^{D \times H \times W}$ denote the ground-truth label volume and $\hat{y} \in [0,1]^{C \times D \times H \times W}$ the predicted softmax probabilities.
The standard nnU-Net loss is $\mathcal{L}_{\text{global}} = \mathcal{L}_{\text{DC+CE}}(\hat{y}, y)$, the sum of soft Dice and cross-entropy losses, computed independently for each foreground class and averaged uniformly.
Under this formulation, each voxel contributes equally to the gradient, so large instances dominate the training signal and small or rare instances are under-represented.

\subsection{Multi-Class Extension of Instance-Sensitive Losses}
Blob loss~\cite{kofler2023} and CC loss~\cite{bouteille2025learning} are instance-aware losses developed for single-class (binary) segmentation.
Their key idea is to decompose the ground-truth label map into connected components, treating each as a separate instance, and computing an independent loss term per component so that every instance contributes equally to the gradient regardless of its size.
However, both losses were designed for single-class segmentation and cannot be applied directly to a multi-class setting with $C{-}1$ foreground classes.

We extend these losses to multi-class segmentation using a one-vs-rest decomposition strategy.
For each foreground class $c \in \{1, \ldots, C{-}1\}$, we extract the binary label map $y_c$ and predicted probability map $\hat{y}_c$ by isolating class $c$ from the multi-class labels (all other foreground classes are treated as background).
We then compute 3D connected components on $y_c$ to obtain the set of instances $\{S_1^c, \ldots, S_{K_c}^c\}$, where $K_c$ is the number of connected components for class $c$.
Each instance is processed independently with the original single-class loss, and results are aggregated uniformly across instances and classes (see Sec.~\ref{sec:aggregation}).

\subsection{Voronoi-Partitioned CC Loss~\cite{bouteille2025learning}}
CC loss partitions the entire volume into non-overlapping regions using a Voronoi tessellation, so that every voxel (including background) is assigned to its nearest component based on the Euclidean distance transform.
Let $V_k^c$ denote the Voronoi cell of component $k$ in class $c$.
The per-component loss evaluates Dice and cross-entropy only within $V_k^c$, restricting both predictions and labels to this spatial region.
This ensures that each component's loss is computed within a local context proportional to its spatial extent, preventing large nearby components from dominating the gradient of smaller ones:
\begin{equation}
	\ell_k^c = \mathcal{L}_{\text{DC+CE}}(\hat{y}_c|_{V_k^c},\; y_c|_{V_k^c}).
	\label{eq:cc}
\end{equation}

\subsection{Blob Loss~\cite{kofler2023}}
Blob loss takes a different approach to isolating each instance.
Rather than partitioning the volume, it constructs an instance-specific domain $\Omega_k^c$ by removing all other foreground components of the same class from the image domain $\Omega$, leaving only the $k$-th component and the background:
\begin{equation}
	\Omega_k^c = \Omega \setminus \bigcup_{j=1,\, j \neq k}^{K_c} S_j^c,
	\label{eq:blob_domain}
\end{equation}
\begin{equation}
	\ell_k^c = \mathcal{L}_{\text{DC+CE}}(\hat{y}_c|_{\Omega_k^c},\; y_c|_{\Omega_k^c}).
	\label{eq:blob}
\end{equation}

\subsection{Instance Loss Aggregation}\label{sec:aggregation}
The instance-level loss averages uniformly over all components and classes:
\begin{equation}
	\mathcal{L}_{\text{instance}} = \frac{1}{|\mathcal{C}|} \sum_{c \in \mathcal{C}} \frac{1}{K_c} \sum_{k=1}^{K_c} \ell_k^{c},
	\label{eq:instance}
\end{equation}
where $\mathcal{C} = \{c : K_c > 0\}$ is the set of classes with at least one instance.
This uniform averaging is where instance imbalance resolution becomes class imbalance resolution: ET with 1{,}739 components and RC with 32 components each receive weight $\frac{1}{|\mathcal{C}|}$ in the outer sum, while the inner $\frac{1}{K_c}$ ensures equal per-instance contribution within each class.
A rare class with few instances is no longer drowned out by common classes with many instances.

\subsection{Inverse-Size Weighting Within Instance Losses}
An alternative approach to instance imbalance is inverse-size weighting.
Shirokikh et al.~\cite{shirokikh2020} proposed this as a standalone global mechanism: each voxel receives weight $w = N / ((K{+}1) \cdot |C_i|)$, where $N$ is total voxels, $K$ is the number of components in the entire volume, and $|C_i|$ is the size (number of voxels) of the component that the voxel belongs to.
The intuition is that voxels belonging to small components receive higher weights, amplifying their contribution to the gradient.
However, when applied globally to multi-class segmentation, the weight magnitudes vary by orders of magnitude across classes (a small component in a rare class can receive a weight thousands of times larger than a large component in a common class), destabilizing training (Tables~\ref{tab:dsc}--\ref{tab:main}, last three rows).

\begin{table*}[t]
	\caption{Region-wise DSC on BraTS-METS 2025 (260 cases). WT = whole tumor, TC = tumor core, ET = enhancing tumor, RC = resection cavity, FG = foreground mean. Best in \textbf{bold}, second-best \underline{underlined}. $\dagger$: $\beta = 2$.}\label{tab:dsc}
	\centering
	\begin{tabular}{lccccc}
		\toprule
		Method                       & WT & TC & ET & RC & FG \\
		\midrule
		Baseline (DC+CE)             & $.751 \pm .28$ & $.749 \pm .28$ & $.727 \pm .27$ & $.150 \pm .30$ & $.594 \pm .27$ \\
		\midrule
		Global + Blob                & $.763 \pm .26$ & $.750 \pm .27$ & $.725 \pm .26$ & $.221 \pm .35$ & $.615 \pm .26$ \\
		Global + Blob$^\dagger$      & $.747 \pm .26$ & $.731 \pm .26$ & $.709 \pm .25$ & $.241 \pm .35$ & $.607 \pm .25$ \\
		Global + CC                  & $\underline{.768} \pm .26$ & $\bm{.764} \pm .27$ & $\bm{.740} \pm .25$ & $.306 \pm .37$ & $\underline{.644} \pm .26$ \\
		Global + CC$^\dagger$        & $\bm{.772} \pm .25$ & $\underline{.762} \pm .27$ & $\underline{.739} \pm .25$ & $.278 \pm .36$ & $.638 \pm .25$ \\
		\midrule
		InvWeight (global)           & $.609 \pm .31$ & $.615 \pm .32$ & $.577 \pm .30$ & $.220 \pm .32$ & $.505 \pm .29$ \\
		InvWeight$^\dagger$ (global) & $.555 \pm .32$ & $.543 \pm .33$ & $.501 \pm .30$ & $.158 \pm .28$ & $.439 \pm .30$ \\
		InvWeight (local)            & $.486 \pm .32$ & $.427 \pm .31$ & $.381 \pm .27$ & $.112 \pm .23$ & $.352 \pm .28$ \\
		\midrule
		Global + IWL-Blob            & $.725 \pm .28$ & $.722 \pm .29$ & $.695 \pm .27$ & $\bm{.441} \pm .36$ & $\bm{.646} \pm .27$ \\
		Global + IWL-Blob$^\dagger$  & $.684 \pm .29$ & $.692 \pm .30$ & $.662 \pm .28$ & $.336 \pm .35$ & $.593 \pm .28$ \\
		Global + IWL-CC              & $.704 \pm .29$ & $.709 \pm .30$ & $.677 \pm .27$ & $\underline{.348} \pm .36$ & $.609 \pm .28$ \\
		Global + IWL-CC$^\dagger$    & $.652 \pm .30$ & $.662 \pm .31$ & $.628 \pm .29$ & $.215 \pm .32$ & $.539 \pm .29$ \\
		\bottomrule
	\end{tabular}
\end{table*}

\begin{table*}[t]
	\caption{Region-wise $PQ_{0.5}^{\text{DSC}}$ on BraTS-METS 2025 (260 cases). Best in \textbf{bold}, second-best \underline{underlined}. $\dagger$: $\beta = 2$.}\label{tab:main}
	\centering
	\begin{tabular}{lccccc}
		\toprule
		Method                       & WT & TC & ET & RC & FG \\
		\midrule
		Baseline (DC+CE)             & $\underline{.466} \pm .27$ & $\underline{.503} \pm .27$ & $\bm{.476} \pm .26$ & $.080 \pm .20$ & $.381 \pm .25$ \\
		\midrule
		Global + Blob                & $\bm{.477} \pm .26$ & $\bm{.509} \pm .27$ & $\underline{.470} \pm .26$ & $.149 \pm .27$ & $\bm{.401} \pm .24$ \\
		Global + Blob$^\dagger$      & $.448 \pm .26$ & $.475 \pm .27$ & $.444 \pm .26$ & $.156 \pm .27$ & $.381 \pm .24$ \\
		Global + CC                  & $.439 \pm .25$ & $.480 \pm .27$ & $.450 \pm .25$ & $\underline{.190} \pm .29$ & $\underline{.390} \pm .24$ \\
		Global + CC$^\dagger$        & $.447 \pm .25$ & $.481 \pm .27$ & $.453 \pm .25$ & $.162 \pm .27$ & $.386 \pm .23$ \\
		\midrule
		InvWeight (global)           & $.118 \pm .15$ & $.171 \pm .18$ & $.147 \pm .16$ & $.083 \pm .19$ & $.129 \pm .16$ \\
		InvWeight$^\dagger$ (global) & $.052 \pm .07$ & $.084 \pm .10$ & $.070 \pm .08$ & $.043 \pm .14$ & $.062 \pm .08$ \\
		InvWeight (local)            & $.010 \pm .02$ & $.017 \pm .02$ & $.014 \pm .02$ & $.017 \pm .07$ & $.015 \pm .02$ \\
		\midrule
		Global + IWL-Blob            & $.291 \pm .22$ & $.380 \pm .25$ & $.350 \pm .24$ & $\bm{.204} \pm .26$ & $.306 \pm .22$ \\
		Global + IWL-Blob$^\dagger$  & $.204 \pm .19$ & $.306 \pm .24$ & $.279 \pm .22$ & $.145 \pm .22$ & $.234 \pm .20$ \\
		Global + IWL-CC              & $.246 \pm .23$ & $.349 \pm .26$ & $.317 \pm .24$ & $.152 \pm .24$ & $.266 \pm .22$ \\
		Global + IWL-CC$^\dagger$    & $.158 \pm .18$ & $.247 \pm .23$ & $.219 \pm .20$ & $.086 \pm .21$ & $.177 \pm .19$ \\
		\bottomrule
	\end{tabular}
\end{table*}

We instead integrate this weighting \emph{within} the per-component instance loss by applying component-specific weights only to voxels within each component's region. For each component $k$ in class $c$, we compute a weight map $w_k^c$ that is applied to both the Dice and cross-entropy terms:
\begin{equation}
	\ell_k^c = \mathcal{L}_{\text{DC+CE}}(w_k^c \odot \hat{y}_c|_{\Omega_k^c},\, w_k^c \odot y_c|_{\Omega_k^c}),
	\label{eq:iwl}
\end{equation}
where $\odot$ denotes element-wise multiplication and $w_k^c(v) = \min(\max(|\Omega_k^c| / |S_k^c|, 1), 2 \times 10^5)$ for voxels $v$ in the component domain $\Omega_k^c$ or $V_k^c$, where $|\Omega_k^c|$ is the number of voxels in the component's domain and $|S_k^c|$ is the number of foreground voxels in component $k$, bounded to $[1, 2 \times 10^5]$ to prevent extreme values.
This confines the reweighting to each component's spatial context, avoiding the training degradation that occurs when weights vary by orders of magnitude across the entire volume.

\begin{table*}[t]
	\caption{Foreground-mean panoptic quality (PQ) at three DSC thresholds $\tau$. Higher is better. Best in \textbf{bold}, second-best \underline{underlined}. $\dagger$: $\beta = 2$.}\label{tab:pq}
	\centering
	\begin{tabular}{lccc}
		\toprule
		Method                       & $\tau = 10^{-6}$ & $\tau = 0.25$ & $\tau = 0.5$ \\
		\midrule
		Baseline                     & $0.412 \pm 0.240$ & $0.408 \pm 0.242$ & $0.381 \pm 0.249$ \\
		Global + Blob                & $\bm{0.436 \pm 0.228}$ & $\bm{0.431 \pm 0.232}$ & $\bm{0.401 \pm 0.240}$ \\
		Global + Blob$^\dagger$      & $0.419 \pm 0.225$ & $0.414 \pm 0.227$ & $0.381 \pm 0.243$ \\
		Global + CC                  & $\underline{0.427 \pm 0.220}$ & $\underline{0.422 \pm 0.224}$ & $\underline{0.390 \pm 0.236}$ \\
		Global + CC$^\dagger$        & $0.422 \pm 0.216$ & $0.416 \pm 0.220$ & $0.386 \pm 0.234$ \\
		\midrule
		InvWeight (global)           & $0.165 \pm 0.159$ & $0.158 \pm 0.158$ & $0.129 \pm 0.156$ \\
		InvWeight$^\dagger$ (global) & $0.083 \pm 0.090$ & $0.078 \pm 0.089$ & $0.062 \pm 0.083$ \\
		InvWeight (local)            & $0.024 \pm 0.026$ & $0.022 \pm 0.025$ & $0.015 \pm 0.021$ \\
		\midrule
		Global + IWL-Blob            & $0.343 \pm 0.218$ & $0.338 \pm 0.218$ & $0.306 \pm 0.216$ \\
		Global + IWL-Blob$^\dagger$  & $0.272 \pm 0.207$ & $0.267 \pm 0.206$ & $0.234 \pm 0.200$ \\
		Global + IWL-CC              & $0.307 \pm 0.223$ & $0.302 \pm 0.224$ & $0.266 \pm 0.224$ \\
		Global + IWL-CC$^\dagger$    & $0.220 \pm 0.194$ & $0.214 \pm 0.194$ & $0.177 \pm 0.189$ \\
		\bottomrule
	\end{tabular}
\end{table*}

\begin{table*}[t]
	\caption{Foreground-mean recognition quality (RQ) at three DSC thresholds $\tau$. Higher is better. Best in \textbf{bold}, second-best \underline{underlined}. $\dagger$: $\beta = 2$.}\label{tab:rq}
	\centering
	\begin{tabular}{lccc}
		\toprule
		Method                       & $\tau = 10^{-6}$ & $\tau = 0.25$ & $\tau = 0.5$ \\
		\midrule
		Baseline                     & $0.598 \pm 0.286$ & $0.562 \pm 0.295$ & $0.494 \pm 0.301$ \\
		Global + Blob                & $\bm{0.651 \pm 0.265}$ & $\bm{0.605 \pm 0.287}$ & $\bm{0.531 \pm 0.292}$ \\
		Global + Blob$^\dagger$      & $0.636 \pm 0.270$ & $\underline{0.599 \pm 0.281}$ & $0.515 \pm 0.302$ \\
		Global + CC                  & $\underline{0.639 \pm 0.256}$ & $0.597 \pm 0.278$ & $\underline{0.516 \pm 0.287}$ \\
		Global + CC$^\dagger$        & $0.635 \pm 0.250$ & $0.587 \pm 0.273$ & $0.509 \pm 0.286$ \\
		\midrule
		InvWeight (global)           & $0.303 \pm 0.241$ & $0.255 \pm 0.224$ & $0.180 \pm 0.205$ \\
		InvWeight$^\dagger$ (global) & $0.164 \pm 0.156$ & $0.127 \pm 0.134$ & $0.088 \pm 0.113$ \\
		InvWeight (local)            & $0.058 \pm 0.059$ & $0.043 \pm 0.048$ & $0.022 \pm 0.032$ \\
		\midrule
		Global + IWL-Blob            & $0.543 \pm 0.295$ & $0.498 \pm 0.298$ & $0.418 \pm 0.277$ \\
		Global + IWL-Blob$^\dagger$  & $0.446 \pm 0.304$ & $0.412 \pm 0.294$ & $0.326 \pm 0.263$ \\
		Global + IWL-CC              & $0.501 \pm 0.301$ & $0.457 \pm 0.297$ & $0.365 \pm 0.281$ \\
		Global + IWL-CC$^\dagger$    & $0.378 \pm 0.286$ & $0.342 \pm 0.273$ & $0.245 \pm 0.242$ \\
		\bottomrule
	\end{tabular}
\end{table*}

\subsection{Combined Loss}
We combine the global and instance losses rather than replacing one with the other:
\begin{equation}
	\mathcal{L} = \alpha \cdot \mathcal{L}_{\text{global}} + \beta \cdot \mathcal{L}_{\text{instance}}.
	\label{eq:combined}
\end{equation}
Retaining the global loss is important because the instance loss, by construction, evaluates each component in isolation and has no mechanism to penalize false positives that fall outside any ground-truth component's region.

The global DC+CE loss operates over the full volume and provides this complementary signal, penalizing spurious predictions in background regions.
Without it, the model would receive no gradient for false positive voxels that do not overlap with any ground-truth instance.
The combination thus balances two objectives: the global loss maintains overall segmentation accuracy and suppresses false positives, while the instance loss ensures that small and rare-class instances are not ignored.
We follow the original blob loss formulation~\cite{kofler2023} and evaluate $\alpha{=}\beta{=}1$ ($1\times$) as well as the recommended $\alpha{=}1, \beta{=}2$ ($2\times$).

\subsection{Evaluation Metrics}
Panoptic quality~\cite{kirillov2019} (PQ) decomposes instance-level evaluation into recognition quality (RQ) and segmentation quality (SQ). At a DSC matching threshold $\tau$, instances are matched using the Hungarian algorithm, yielding true positives (TP), false positives (FP), and false negatives (FN):
\begin{equation}
	\text{PQ}_\tau^{\text{DSC}} = \underbrace{\frac{|\text{TP}|}{|\text{TP}| + \frac{1}{2}|\text{FP}| + \frac{1}{2}|\text{FN}|}}_{\text{RQ}} \times \underbrace{\frac{1}{|\text{TP}|} \sum_{(p,g) \in \text{TP}} \text{DSC}(p, g)}_{\text{SQ}},
	\label{eq:pq}
\end{equation}
where $p$ and $g$ denote predicted and ground truth instances, respectively, and $\tau \in \{10^{-6}, 0.25, 0.5\}$ is the DSC threshold for instance matching. We substitute DSC for IoU in the SQ term, as DSC is the standard overlap metric in medical image segmentation; reported $PQ_\tau^{\text{DSC}}$ values are therefore not directly comparable to IoU-based PQ. We prefer $PQ_\tau^{\text{DSC}}$ over alternative metrics like CC-Metrics~\cite{jaus2025every} because: (1) the matching threshold $\tau$ is tunable, allowing assessment across multiple operating points; (2) CC-Metrics relies on the same Voronoi decomposition as CC loss, which could introduce evaluation bias in favor of that loss.

\begin{figure*}[t]
	\centering
	\includegraphics[width=0.7\textwidth]{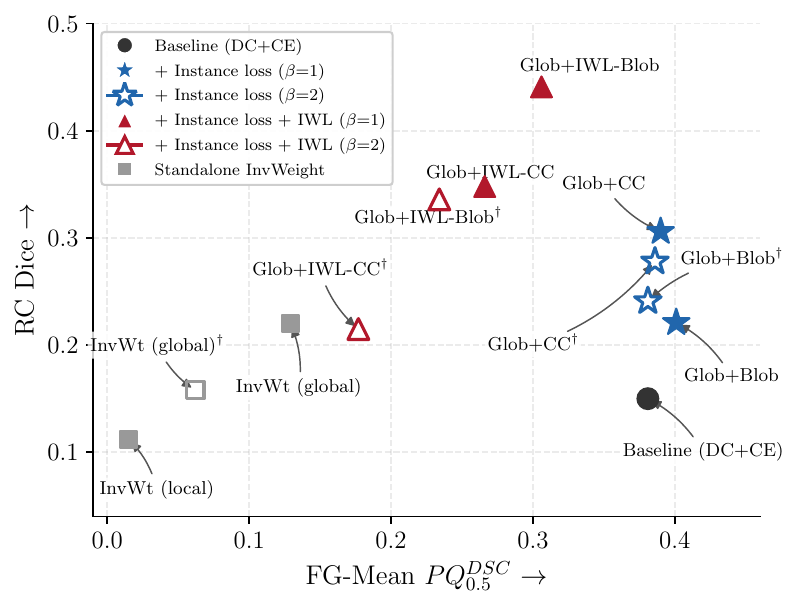}
	\caption{RC Dice vs.\ FG-Mean $PQ_{0.5}^{\text{DSC}}$. Filled markers: $\beta{=}1$; hollow: $\beta{=}2$. Instance losses (stars) improve RC Dice over the baseline while preserving $PQ_{0.5}^{\text{DSC}}$. IWL variants (triangles) further increase RC Dice at the cost of $PQ_{0.5}^{\text{DSC}}$. Standalone inverse weighting (squares) shows low performance on both metrics.}\label{fig:tradeoff}
\end{figure*}

\section{Experiments}

\subsection{Dataset and Evaluation}
BraTS-METS 2025~\cite{maleki2025analysis} comprises 1{,}295 post-treatment brain MRI cases (1{,}035 train, 260 test). The challenge uses hierarchical labels: WT = labels $\{1,2,3\}$, TC = labels $\{1,3\}$, ET = label $\{3\}$, RC = label $\{4\}$, where labels 1 and 2 correspond to Non-enhancing tumor core (NETC) and Surrounding non-enhancing FLAIR hyperintensity (SNFH), respectively.

\subsection{Metric Decisions}
We report region-wise DSC and multi-threshold $PQ_{\tau}^{\text{DSC}}$ (Eq.~\eqref{eq:pq}) at DSC thresholds $\tau \in \{10^{-6}, 0.25, 0.5\}$, plus size-stratified $PQ_{0.5}^{\text{DSC}}$.

\subsection{Training}
All methods use nnU-Net 3D full-resolution~\cite{isensee2021} trained for 1{,}000 epochs on the BraTS-METS 2025 training set (1{,}035 cases). Preprocessing, patch size, batch size, augmentation, and optimizer follow the nnU-Net defaults. The only modification across methods is the loss function. Reported standard deviations reflect patient-level variability across the 260 test cases.

\section{Results}

\subsection{Reweighting}
\emph{Where you reweight matters more than whether you reweight.}
A recurring theme across our experiments is that the effectiveness of reweighting depends on the scope at which it is applied, not just whether it is applied.
Inverse-size weighting applied globally across the entire volume (InvWeight rows in Tables~\ref{tab:dsc}--\ref{tab:main}) produces the worst results in both DSC and $PQ_{0.5}^{\text{DSC}}$, with foreground Dice as low as 0.352 and $PQ_{0.5}^{\text{DSC}}$ as low as 0.015.
The same reweighting principle, when confined to the spatial context of each connected component (IWL-Blob, IWL-CC), yields the best RC Dice (0.441) and competitive foreground Dice (0.646).
The difference is not the weighting scheme itself but its scope: global application creates weight imbalances across classes (a small component in a rare class can receive a weight orders of magnitude larger than a large component in a common class), destabilizing optimization.
Local application within per-component losses bounds the weight variation to a single component's region, where the ratio between foreground and background voxels is controlled.
Similarly, the one-vs-rest class decomposition in blob and CC loss is itself a form of structured reweighting: by averaging uniformly over classes, it ensures each class contributes equally without requiring explicit class weights.
This suggests that structured, spatially scoped reweighting is more effective than uniform global reweighting for multi-class segmentation with imbalanced classes and instances.

\begin{figure*}[t]
	\centering
	\includegraphics[width=\textwidth]{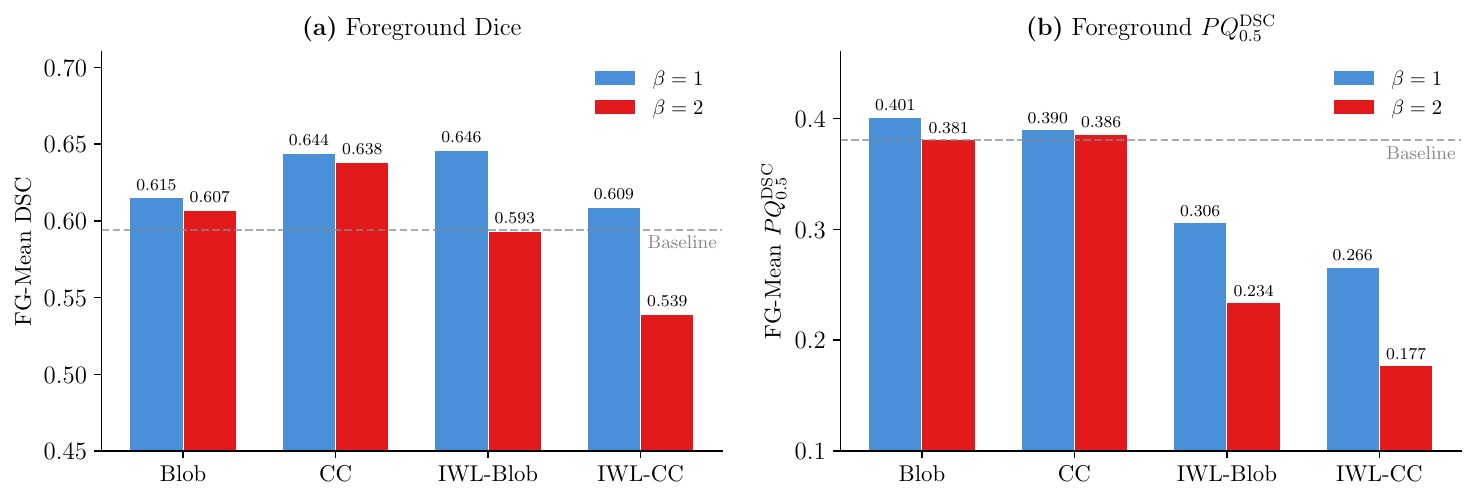}
	\caption{Effect of $\beta$ weighting on foreground-mean DSC and $PQ_{0.5}^{\text{DSC}}$. Dashed line: baseline (DC+CE). $\beta{=}1$ (blue) consistently outperforms $\beta{=}2$ (red) across all methods, with the gap widening for IWL variants.}\label{fig:beta}
\end{figure*}

\subsection{Voxel-Level Performance}
\emph{Instance-aware losses improve rare-class Dice without sacrificing common-region performance.}
Table~\ref{tab:dsc} reports region-wise DSC across all methods.
The baseline DC+CE achieves strong performance on the three common regions (WT 0.751, TC 0.749, ET 0.727) but fails on the rare resection cavity (RC 0.150).
Multi-class CC loss improves DSC across all four regions simultaneously, achieving the best TC (0.764) and ET (0.740) and raising RC from 0.150 to 0.306, yielding the second-best foreground mean (0.644).
Blob loss shows a similar but smaller improvement (RC 0.221, FG 0.615), consistent with single-class findings where CC loss outperforms blob loss on voxel overlap~\cite{bouteille2025learning}.

Integrating inverse-size weighting within the instance loss (IWL-Blob) produces the highest RC Dice (0.441) and the best foreground mean (0.646), but at the cost of reduced performance on common regions (WT drops from 0.768 to 0.725, ET from 0.740 to 0.695).
Standalone inverse-size weighting (last three rows) degrades performance across all regions, confirming that global application is not viable in the multi-class setting.

\subsection{Instance-Level Performance}
\emph{Blob loss achieves the best detection quality; CC loss follows closely with better voxel overlap.}
Foreground-mean $\text{PQ}$ and $\text{RQ}$ at three DSC thresholds are shown in Tables~\ref{tab:pq} and~\ref{tab:rq}.
Global + Blob achieves the highest $\text{PQ}$ across all thresholds ($PQ_{0.5}^{\text{DSC}} = 0.401 \pm 0.240$, $\text{RQ} = 0.531 \pm 0.292$), followed by Global + CC ($0.390 \pm 0.236$, $0.516 \pm 0.287$).
The $\beta{=}2$ variants consistently underperform their $\beta{=}1$ counterparts, in contrast to the single-class setting where $\beta{=}2$ was recommended~\cite{kofler2023}.
IWL variants achieve lower $PQ_{0.5}^{\text{DSC}}$ despite higher Dice (e.g., IWL-Blob: $0.306 \pm 0.216$), revealing a systematic Dice--$PQ_{0.5}^{\text{DSC}}$ trade-off.

\subsection{The Dice--$PQ_{0.5}^{\text{DSC}}$ Trade-off}
\emph{Improving rare-class overlap comes at the cost of instance-level detection.}
Figure~\ref{fig:tradeoff} plots RC Dice against $PQ_{0.5}^{\text{DSC}}$, showing three clusters: Blob/CC ($PQ_{0.5}^{\text{DSC}}$ 0.39--0.40, RC Dice 0.22--0.31), IWL (RC Dice 0.22--0.44, $PQ_{0.5}^{\text{DSC}}$ 0.18--0.31), and standalone inverse weighting (both metrics below 0.13).
Table~\ref{tab:size} stratifies $PQ_{0.5}^{\text{DSC}}$ by instance size.
Blob and CC maintain small- and medium-instance $PQ_{0.5}^{\text{DSC}}$ (0.22--0.61) while trading off large-instance detection quality (0.703 to 0.730).
The IWL variants degrade small-instance $PQ_{0.5}^{\text{DSC}}$ (Blob: $0.241 \pm 0.247$ to $0.153 \pm 0.191$ under IWL-Blob) but maintain large-instance detection, suggesting a shift in the model's detection strategy towards larger instances.

\subsection{Effect of $\beta$ Weighting}
\emph{The multi-class decomposition already amplifies rare classes, making $\beta{=}2$ redundant.}
In single-class segmentation, Kofler et al.~\cite{kofler2023} recommended $\beta{=}2$ (doubling the instance loss weight relative to the global loss).
In our multi-class setting, $\beta{=}1$ consistently outperforms $\beta{=}2$ across all configurations (Fig.~\ref{fig:beta}).
For CC loss, increasing $\beta$ from 1 to 2 reduces foreground Dice from 0.644 to 0.638 and $PQ_{0.5}^{\text{DSC}}$ from 0.390 to 0.386.
The effect is more pronounced for IWL variants: IWL-Blob drops from 0.646 to 0.593 in foreground Dice and from 0.306 to 0.234 in $PQ_{0.5}^{\text{DSC}}$.
We attribute this to the class-balancing effect already introduced by the one-vs-rest decomposition.
In single-class segmentation, $\beta{=}2$ compensates for the global loss dominating the instance loss.
In multi-class segmentation, the instance loss already carries more weight per rare-class instance (due to uniform class averaging), so further increasing $\beta$ over-corrects, pushing the model to over-segment rare classes at the expense of common-region accuracy.

\section{Conclusion}

We showed that extending instance-sensitive losses to multi-class segmentation via one-vs-rest decomposition repurposes them to solve class imbalance in addition to instance imbalance, and that inverse-size weighting is effective when integrated within per-component losses rather than applied globally.
On BraTS-METS 2025, multi-class CC loss raises RC Dice from 0.15 to 0.31 while maintaining $PQ_{0.5}^{\text{DSC}}$, and IWL-Blob further increases it to 0.44 at the cost of lower detection quality.
Our results reveal a systematic trade-off between voxel-level overlap and instance-level detection; it will be interesting to investigate how this trade-off can be overcome.
From a clinical perspective, these improvements in resection cavity segmentation directly benefit post-operative SRS planning, and more broadly, methods that account for both instance and class imbalance can help close the gap between automated segmentation and the per-lesion accuracy required for individualized treatment of brain metastases.

\section{Acknowledgement}
SSK acknowledges support from a grant from the United Kingdom Research and Innovation Medical Research Council awarded as part of a Doctoral Training Partnership at King’s College London (MR/W006820/1). TV Medtronic/RAEng [RCSRF1819$\$7$\$34], and Wellcome/EPSRC
[WT203148/Z/16/Z; NS/A000049/1]


{\small
\bibliographystyle{ieeenat_fullname}
\bibliography{main}
}

\end{document}